\documentclass{article}

\usepackage{PRIMEarxiv}

\usepackage[utf8]{inputenc} 
\usepackage[T1]{fontenc}    
\usepackage{hyperref}       
\usepackage{url}            
\usepackage{booktabs}       
\usepackage{amsfonts}       
\usepackage{nicefrac}       
\usepackage{microtype}      
\usepackage{lipsum}
\usepackage{fancyhdr}       
\usepackage{graphicx}       
\usepackage{tabularx} 
\usepackage{amsthm} 
\usepackage{multirow} 
\usepackage{epsfig}
\usepackage{epstopdf}
\usepackage{graphics}
\usepackage{float} 
\usepackage{wrapfig}
\usepackage{amsmath}
\usepackage[T1]{fontenc}
\usepackage[utf8]{inputenc}
\usepackage{authblk}

\graphicspath{{media/}}     

\title{Rethinking the Number of Shots in\\Robust Model-Agnostic Meta-Learning
}
\author[1]{Xiaoyue Duan}
\author[1]{Guoliang Kang}
\author[1]{Runqi Wang}
\author[2]{Shumin Han}
\author[2]{Song Xue}
\author[1]{Tian Wang}
\author[1]{Baochang Zhang}
\affil[1]{Beihang University, Beijing, China}
\affil[2]{VIS, Baidu, Beijing, China}

\date{} 

\begin{document}
\maketitle

\begin{abstract}
Robust Model-Agnostic Meta-Learning (MAML) is usually adopted to train a meta-model which may fast adapt to novel classes with only a few exemplars and meanwhile 
remain robust to adversarial attacks.
The conventional solution for robust MAML is to introduce robustness-promoting regularization during meta-training stage.
With such a regularization, previous robust MAML methods simply follow the typical MAML practice that the number of training shots should match with the number of test shots to achieve an optimal adaptation performance. 
However, although the robustness can be largely improved, previous methods sacrifice clean accuracy a lot. 
In this paper, we observe that introducing robustness-promoting regularization into MAML reduces the intrinsic dimension of clean sample features, 
which results in a lower capacity of clean representations. 
This may explain why the clean accuracy of previous robust MAML methods drops severely.
Based on this observation, we propose a simple strategy, \emph{i.e.}, increasing the number of training shots, to mitigate the loss of intrinsic dimension caused by robustness-promoting regularization.
Though simple, our method remarkably improves the clean accuracy of MAML without much loss of robustness, producing a robust yet accurate model. 
Extensive experiments demonstrate that our method outperforms prior arts in achieving a better trade-off between accuracy and robustness.
Besides, we observe that our method is less sensitive to the 
number of fine-tuning steps during meta-training, which allows for a reduced number of fine-tuning steps to improve training efficiency.
\end{abstract}


\section{Introduction}
\label{sec:intro}

Few-shot learning ~\cite{finn2017model, rajasegaran2020self, li2021learning, dong2022improving} aims to train a model which can fast adapt to novel classes with only a few exemplars. Model-agnostic meta-learning (MAML)~\cite{finn2017model} is a typical meta-learning
approach to deal with few-shot learning problems. However, the model trained through MAML is not robust to adversarial attacks. The conventional adversarial training can facilitate MAML with adversarial robustness. 
However, the limited data in the few-shot setting makes it challenging~\cite{goldblum2020adversarially} to keep both clean accuracy and robustness at a high level at the same time.

\begin{wrapfigure}[29]{r}{0.45\linewidth}
    \begin{center}
        \includegraphics[width=0.95\linewidth]{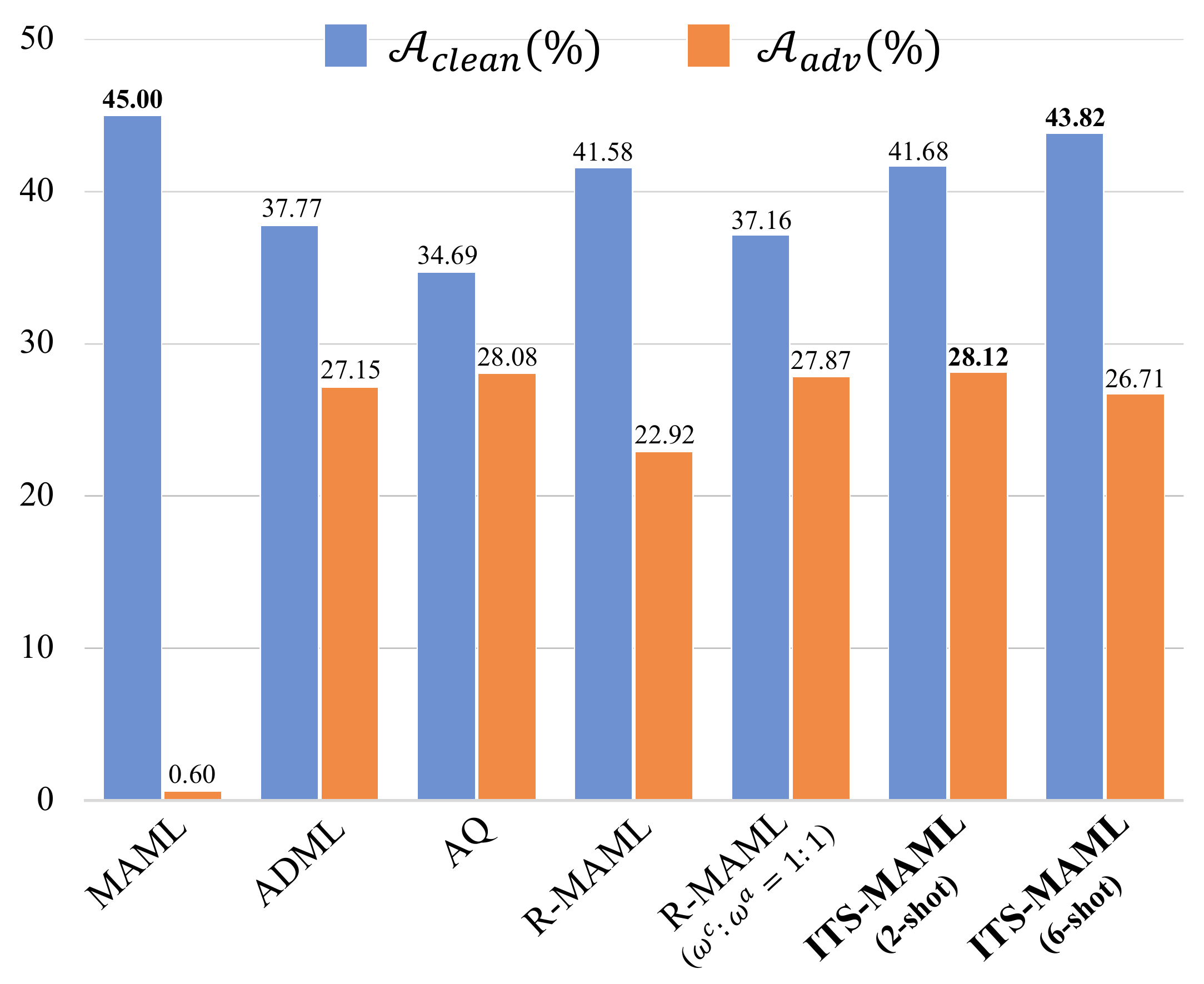}
    \end{center}
    \caption{Accuracy of models trained by MAML, ADML, AQ, R-MAML and our ITS-MAML on 5-way 1-shot miniImageNet~\cite{vinyals2016matching} task. The $\mathcal{A}_{clean}$ and $\mathcal{A}_{adv}$  denote clean accuracy and robust accuracy respectively. We adopt a 10-step PGD attack~\cite{madry2017towards} with power $\epsilon=2$. The $w^c$ and $w^a$ are the weights of clean loss and adversarial loss, respectively. The original setting of R-MAML is approximately equivalent to  $w^c\!:\!w^a\!=\!1\!:\!0.2$ for higher clean accuracy. We also report the results of R-MAML with $w^c\!:\!w^a\!=\!1\!:\!1$ for a fair comparison with other methods.}
    \label{fig:motivation_trade_off_figure}
\end{wrapfigure}

In recent years, a series of works~\cite{yin2018adversarial, goldblum2020adversarially, wang2021fast} paid attention to the robust MAML. 
Most of them attempt to introduce robustness-promoting regularization (\emph{i.e.}, adversarial loss) into the typical MAML bi-level training framework, \emph{e.g.},  AQ~\cite{goldblum2020adversarially}, ADML~\cite{yin2018adversarial}, and R-MAML~\cite{wang2021fast}.
Although those methods introduce robustness-promoting regularization in different ways, 
all of them follow the typical MAML training principal that the number of training shots should match with the number of test shots to 
achieve the best novel class adaptation performance. 
We fairly compare the performance of those methods (see Fig.~\ref{fig:motivation_trade_off_figure})
and find that 
compared to plain MAML, all of those robust MAML methods greatly sacrifice the clean accuracy to improve their robustness. 
Thus, a natural question arises: 
what is the underlying factor that causes the severe loss of clean accuracy?

In deep CNNs, intrinsic dimension of feature embedding, which is defined as the minimum variations captured to describe a representation or realize a specific optimization goal, has proved to be an accurate predictor of the network’s classification accuracy on the test set ~\cite{ansuini2019intrinsic,cao2019theoretical,aghajanyan2020intrinsic}. Through the lens of intrinsic dimension, we observe an interesting phenomenon existing in robust MAML methods. 
As shown in Table~\ref{tab: intrinsic dimension full}, it can be seen that given the same number of training shots, 
introducing robustness-promoting regularization into MAML framework results in a lower intrinsic dimension of clean feature embedding than the plain MAML framework without adding any adversarial examples or adversarial loss into training. 
Recall that the optimization objective in plain MAML framework is to maximize the likelihood on clean query images, which can be roughly viewed as maximizing the generalization ability to unseen data in each episode. 
Thus, the intrinsic dimension of feature embedding in plain MAML actually represents the minimum variations needed to guarantee the generalization ability in each few-shot task. 
This may explain why introducing adversarial loss into MAML framework may obviously sacrifice the clean accuracy, as introducing 
adversarial loss reduces the required intrinsic dimension, thus resulting in a lower capacity of clean representations.
This phenomenon implies that the conventional MAML practice may not be optimal to robust MAML 
when we care about both the robustness and clean accuracy.

\begin{table}[t]
\centering
\caption{Intrinsic dimensions of features trained by plain MAML and the robust MAML~\cite{wang2021fast} with different numbers of training shots. The intrinsic dimension is estimated through principal component analysis (PCA) of features trained on miniImageNet~\cite{fan2010intrinsic}. 
In our experiments, the intrinsic dimension is set to the number of principal components which retain at least $90\%$ of the variance.
The ``C'' and ``N'' denote the intrinsic dimension of clean samples and the adversarial noise (added to clean samples), respectively.}
\label{tab: intrinsic dimension full}
\setlength{\tabcolsep}{2mm}{
\begin{tabular}{ccccccccc}
\toprule[1pt]
\multicolumn{2}{c}{\multirow{2}{*}{Method}} & \multicolumn{7}{c}{Number of training shots} \\ \cmidrule[0.3pt]{3-9} 
\multicolumn{2}{c}{}                        & 1   & 2    & 3    & 4    & 5    & 6    & 7   \\ \midrule[1pt]
\multirow{1}{*}{MAML}       & C         & 80  & 103  & 139  & 151  & 157  & 160  & 179 \\\midrule[0.3pt]
\multirow{2}{*}{Robust MAML}   & C         & 22  & 41   & 59   & 60   & 71   & 86   & 89  \\
                            & N         & 4   & 7    & 7    & 8    & 8    & 9    & 10  \\ \bottomrule[1pt]
\end{tabular}}
\end{table}


From Table~\ref{tab: intrinsic dimension full}, we also observe that 
with increasing the number of training shots, the intrinsic dimension of clean representations also increases for both
plain MAML and robust MAML framework. 
In this paper, based on our observations, we propose a simple way, \emph{i.e.}, \textit{increasing the number of training shots},
to mitigate the loss of intrinsic dimension caused by robustness-promoting regularization.
For example, if we consider the $N$-way $1$-shot task, we may use a larger number of training shots in robust MAML, \emph{i.e.} larger than 1.
With this simple operation, our method improves clean accuracy of robust MAML remarkably without much loss of robustness. 
Extensive experiments on miniImageNet~\cite{vinyals2016matching}, CIFAR-FS~\cite{bertinetto2018meta}, and Omniglot~\cite{lake2015human} demonstrate that our method performs favourably against previous robust MAML methods considering both clean accuracy and robustness. We also show that our method achieves a better trade-off between clean accuracy and robustness. 
Finally, we demonstrate that compared to previous robust MAML methods, our method is less sensitive to the number of fine-tuning (inner-loop) steps in meta-training, and bares almost no drop of accuracy even when the number of fine-tuning steps is greatly reduced, thus improving training efficiency. 

In a nutshell, our contributions are summarized as: 
\begin{itemize}
\item
We observe that introducing robustness-promoting regularization into MAML reduces the intrinsic dimension of features, which 
means the capacity of representations may be largely reduced. This implies the conventional MAML practice that the number of training shots should match with the number of test shots is not optimal for robust MAML setting.
\item
Based on our observations, we propose a simple yet effective strategy, \emph{i.e.}, increasing the number of training
shots, to mitigate the loss of intrinsic dimension caused by robustness-promoting regularization.
\item
Extensive experiments on three few-shot learning benchmarks, \emph{i.e.}, miniImageNet~\cite{vinyals2016matching}, CIFAR-FS~\cite{bertinetto2018meta}, and Omniglot~\cite{lake2015human}, demonstrate that our method performs favourably against previous robust MAML methods considering both clean accuracy and robustness.
We also demonstrate that our method can achieve a better trade-off between clean accuracy and robustness and may 
improve the training efficiency by reducing the fine-tuning steps at the meta-training stage. 
\end{itemize}

\section{Related work}
\label{sec:related}


\textbf{Adversarial robustness.} Adversarial robustness refers to the accuracy of the model on adversarially perturbed samples, which are visually indistinguishable from clean samples but can drastically change the model predictions~\cite{ilyas2019adversarial, xie2019intriguing}. Adversarial training is one of the most effective approaches to improve the model's adversarial robustness~\cite{goodfellow2014explaining, madry2017towards, zhang2019theoretically}. For example, \cite{goodfellow2014explaining} proposed for the first time to adopt single-step-based adversarial samples for adversarial training, and \cite{madry2017towards} further extended it to multi-step-based adversarial examples for better robustness. 
Others propose methods for fast adversarial training to improve the training efficiency~\cite{shafahi2019adversarial, zhang2019you, wong2020fast}.
Methods have been proposed to improve the transferablity of robustness in few-shot settings~\cite{chen2020adversarial, chan2020thinks, tian2020rethinking, rizve2021exploring}. However, learning an adversarially robust meta-model is challenging. The improvement of robustness is often accompanied by the decline of accuracy since a fundamental trade-off between the clean and adversarial distributions exists, as is theoretically proved by \cite{zhang2019theoretically}. The scarce data in few-shot settings makes it harder to learn the robust decision boundary, resulting in even more vulnerability of the model against adversarial attacks~\cite{xu2021yet}.

\textbf{Adversarially robust model-agnostic meta-learning.}
Meta-learning has demonstrated promising performance in few-shot learning~\cite{snell2017prototypical, huang2018natural, maicas2018training, sung2018learning, wang2020tracking}. MAML~\cite{finn2017model}, as the first to propose an effective meta-learning framework, can learn a well-initialized meta-model to quickly adapt to new few-shot classification tasks. Though widely adopted, MAML naturally lacks adversarial robustness.
A few work studied the robustness of MAML including ADML~\cite{yin2018adversarial}, AQ~\cite{goldblum2020adversarially} and R-MAML~\cite{wang2021fast}. However, these methods do not achieve a satisfactory trade-off between clean accuracy and adversarial robustness. For example, AQ trades accuracy for robustness compared with ADML, while R-MAML achieves a high clean accuracy on condition that the robustness is greatly reduced compared with AQ (see Fig.~\ref{fig:motivation_trade_off_figure}). In addition, all these methods greatly sacrifice clean accuracy compared with plain MAML.

\section{Methodology}
\label{sec:methodology}


\subsection{Preliminary}
\textbf{MAML for few-shot learning.}
Few-shot learning aims to enable the model to classify data from novel classes with only a few (\emph{e.g.}, 1) examples to train. Few-shot learning is typically formulated as a $N$-way $K$-shot classification problem, where in each task, we aim to classify samples (denoted as ``query'') into $N$ different classes, with $K$ samples (denoted as ``support'') in each class for training. Meta-learning is the conventional way to deal with the few-shot learning problem, while model-agnostic meta-learning (MAML) is one of the most popular and effective meta-learning methods.

Generally speaking, MAML attempts to learn a well-initialized model (\emph{i.e.}, a \textit{meta-model}), which can quickly adapt to new few-shot classification tasks. 
During meta-training, the meta-model is fine-tuned over $N$ classes (with each class containing $K$ samples), and then  updated by minimizing the validation error of the fine-tuned network over unseen samples from these $N$ classes. The \textit{inner fine-tuning} stage and the \textit{outer meta-update} stage form the \textit{bi-level} learning procedure of MAML. 
Formally, we consider $T$ few-shot classification tasks, each of which contains a support data set $\mathcal{S}=\{s_1, s_2, \cdots, s_K\}$ (${s_j}\in\mathcal{S}$ denotes the $j$-th support sample) for fine-tuning, and a query data set $\mathcal{Q}$ for meta-update. MAML's bi-level optimization problem can be described as:
\begin{equation}
\begin{aligned}
    \underset{\mathbf{\theta}}{\operatorname{minimize}}\ \frac{1}{T}\sum_{i=1}^T&\mathcal{L}_{\mathbf{\theta}_i'}(\mathcal{Q}_i), \\
    \operatorname{subject\ to\ }\mathbf{\theta}_i'\!=\!\underset{\mathbf{\theta}}{\arg\min}\ &\mathcal{L}_{\mathbf{\theta}}(\mathcal{S}_i),\ \forall i\!\in\!\{1,2,\dots,T\},
\end{aligned}
\label{equation: MAML}
\end{equation}
where $\mathbf{\theta}$ is the meta-model to be learned, $\mathbf{\theta}_i'$ is the fine-tuned parameters for the $i$-th task, $\mathcal{L}_\theta(\mathcal{S}_i)$ and $\mathcal{L}_{\mathbf{\theta}_i'}(\mathcal{Q}_i)$ are the training error on the support set and the validation error on the query set, respectively. Note that the fine-tuning stage (corresponding to the constraint in Eq.~\ref{equation: MAML}) usually calls for $M$ steps of gradient update:
\begin{equation}
    \mathbf{\theta}_i^{(m)}\!=\!\mathbf{\theta}_i^{(m\!-\!1)}\!-\!\alpha\nabla_{\mathbf{\theta}_i^{(m\!-\!1)}}\mathcal{L}_{\mathbf{\theta}_i^{(m\!-\!1)}}(\mathcal{S}_i),\ m\!\in\!\{1,2,\dots,M\},
\label{equation: fine-tuning steps}
\end{equation}
where $\alpha$ is the learning rate of the inner update, $\mathbf{\theta}_i^{(0)}=\mathbf{\theta}$ and $\mathbf{\theta}_i^{(M)}=\mathbf{\theta}_i'$. 

\begin{figure*}[t]
    \centering
    \includegraphics[width=0.9\linewidth]{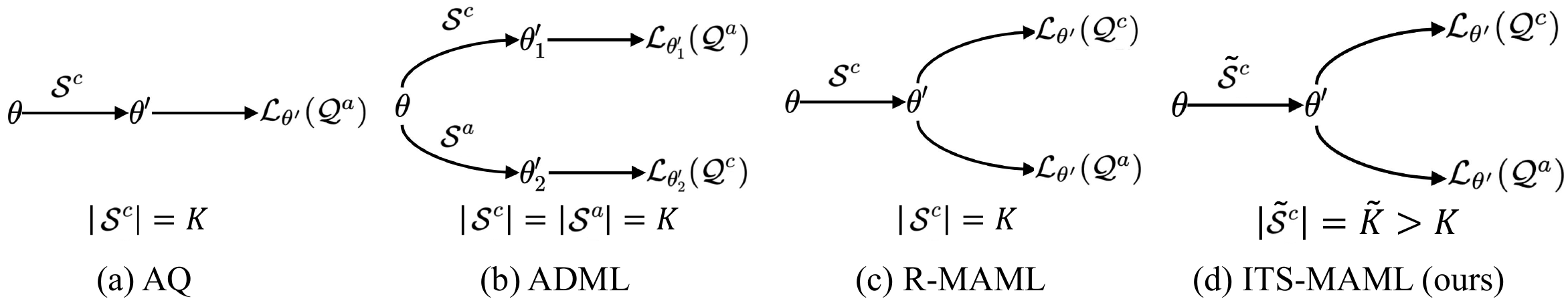}
    \caption{Robust meta-learning frameworks of previous arts and our method. $\mathbf{\theta}$ denotes the initial model parameters in each episode, $\mathbf{\theta}'$ denotes the model parameters after fine-tuning on the support set (inner-loop), and $\mathcal{L}$ is the loss of the fine-tuned model on the query set, adopted for the meta-update of $\mathbf{\theta}$ (outer-loop). $\mathcal{S}$ and $\mathcal{Q}$ are the support data and the query data, with the superscripts $c$ and $a$ denoting clean and adversarial samples respectively. For a $N$-way $K$-shot meta-testing task, our method simply sets the training shot number $\tilde{K}$ larger than $K$.
    Note that each episode consists of several few-shot classification tasks. We omit the task index in our symbols for brevity.}
    \label{fig:previous_frameworks}
\end{figure*}

\textbf{Robust MAML.}
Adversarial training is one of the most effective defense methods to learn a robust model against adversarial attacks~\cite{madry2017towards}. Suppose $\mathcal{D}=\{\mathcal{D}^c, \mathcal{D}^a\}$ denotes the set of samples used for training,
and the adversarial training can be represented as
\begin{equation}
    \underset{\mathbf{\theta}}{\operatorname{minimize}}\quad w^c\cdot\mathcal{L}_\mathbf{\theta}(\mathcal{D}^c)+w^a\cdot\mathcal{G}_\mathbf{\theta}(\mathcal{D}^a),
\end{equation}
where $\mathbf{\theta}$ is the parameters of robust model to be learned, $\mathcal{L}_\mathbf{\theta}(\mathcal{D}^c)$ is the prediction loss (\emph{e.g.}, cross-entropy loss) on clean sample set $\mathcal{D}^c$, $\mathcal{G}_\mathbf{\theta}(\mathcal{D}^a)$ is the adversarial loss (\emph{e.g.}, cross entropy loss~\cite{goodfellow2014explaining, madry2017towards} and KL divergence~\cite{zhang2019theoretically}) on adversarial sample set $\mathcal{D}^a$, and $w^c$ and $w^a$ are the weights of clean loss and adversarial loss respectively. Each adversarial sample $x^a\in\mathcal{D}^a$ is obtained by adding perturbations to the clean sample $x^c\in\mathcal{D}^c$ to maximize its classification loss:
\begin{equation}
    x^a=x^c+\underset{\lVert\mathbf{\delta}\rVert_p\leq\epsilon}{\operatorname{\arg\max}}\ \mathcal{G}_\mathbf{\theta}(x^c+\mathbf{\delta}),
\end{equation}
where the $p$-norm of the perturbation $\mathbf{\delta}$ is limited within the $\epsilon$ bound so that the adversarial samples are visually indistinguishable from the clean samples. 

Existing robust MAML methods introduced adversarial loss to MAML's bi-level learning procedure. 
As illustrated in Fig.~\ref{fig:previous_frameworks}, AQ~\cite{goldblum2020adversarially} directly replaced the loss on clean query images of plain MAML with adversarial loss. 
Compared to AQ, ADML~\cite{yin2018adversarial} additionally added another optimization pathway, \emph{i.e.}, fine-tuning with the 
adversarial loss on support data and evaluating the accuracy on clean query data. 
Further, R-MAML~\cite{wang2021fast} showed that there is no need to impose adversarial loss during the fine-tuning stage and 
imposed both the clean prediction loss and the adversarial loss. 
Although R-MAML demonstrated superior clean accuracy compared to ADML, we find that if we treat the clean loss and adversarial loss on query data equally, \emph{i.e.}, setting the learning rate of clean loss to that of adversarial loss, 
the clean accuracy of R-MAML actually performs comparably to ADML (as shown in Fig.~\ref{fig:motivation_trade_off_figure}).
Thus, R-MAML actually improves clean accuracy by sacrificing robustness (\emph{i.e.}, increasing the learning rate of clean loss).
There remains a question about how to improve clean accuracy while maintaining good robustness.

\begin{figure*}[t]
    \centering
    \includegraphics[width=1.0\linewidth]{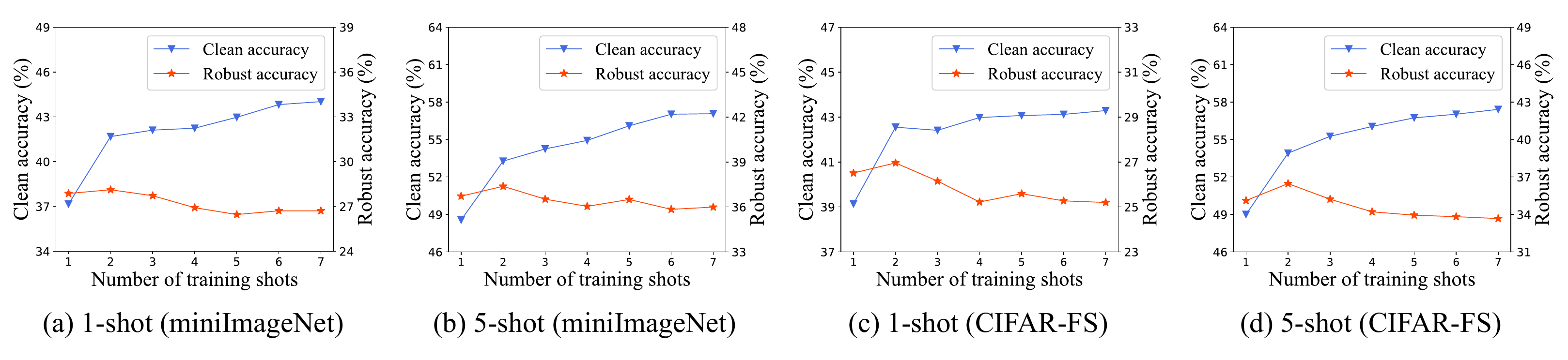}
    \caption{Clean accuracy and robust accuracy of models on 5-way 1-shot and 5-way 5-shot meta-testing tasks. The experiments are conducted on miniImageNet and CIFAR-FS. The models are trained with different numbers of training shots. The robust accuracy is computed with a 10-step PGD attack with $\epsilon\!=\!2$ for miniImageNet and $\epsilon\!=\!8$ for CIFAR-FS.
    }
    \label{fig: acc of ITS-MAML under different test shots}
\end{figure*}

\subsection{Rethinking the number of shots}
\textbf{Robustness regularization reduces intrinsic dimension of features.}
Intrinsic dimension of features denotes the minimum variations captured to describe the representations. From the optimization objective, it can be viewed as the minimum free factors or parameters needed to realize the optimization objective.
Previous works~\cite{ansuini2019intrinsic,cao2019theoretical,aghajanyan2020intrinsic} on intrinsic dimension showed that the intrinsic dimension of features is usually smaller than the length of feature vectors (\emph{e.g.}, the number of units in a specific layer of deep neural network). 
For a $\mathbb{R}^D$ feature space, the intrinsic dimension (denoted as $\hat{D}$) usually satisfies $\hat{D}<D$.
In deep CNN, previous work~\cite{ansuini2019intrinsic} demonstrated that the intrinsic dimension of 
last hidden layer performs as an accurate predictor of network's generalization ability. 
Thus, in this paper, we attempt to analyze the drop of clean accuracy in robust MAML through the lens of intrinsic dimension.

There are different ways to estimate the intrinsic dimension. One of the simplest ways to estimate $\hat{D}$ is to adopt 
Principal Component Analysis (PCA) to discover the number of principal components or variations~\cite{huang2018mechanisms,cao2019theoretical}. 
In this paper, we adopt PCA as our tool to estimate the intrinsic dimension. 
More complex and accurate estimation methods can also be employed~\cite{huang2018mechanisms,facco2017estimating,amsaleg2015estimating}. 
We empirically find that the trends of different estimations are the same regarding the issue that we care about.
More details about how to estimate the intrinsic dimension are presented in Appendix.

As shown in Table~\ref{tab: intrinsic dimension full}, we observe that given the same number of training shots, 
introducing adversarial loss into MAML framework results in a lower intrinsic dimension of clean samples than the plain MAML framework without adding any adversarial loss into training. 
We speculate that it is because unexpected adversarial noise disturbs the normal representation learning on clean data.
Recall that the optimization objective in plain MAML is to maximize the likelihood on clean query images, which can be roughly viewed as maximizing the generalization ability to unseen data in each episode. 
Thus, from the optimization perspective, the intrinsic dimension in plain MAML actually represents the minimum variations needed to guarantee the generalization ability on clean data in each few-shot setting.  
This may explain why introducing adversarial loss into MAML framework may obviously sacrifice the clean accuracy, as introducing 
adversarial loss reduces the required intrinsic dimension and lowers the capacity of clean representations. 

We also investigate how the intrinsic dimension of adversarial noise changes as the training shot number
increases in robust MAML. 
We observe that the intrinsic dimension of adversarial noise for robust MAML is 
insensitive to the number of training shots, indicating that the adversarial robustness may not be affected by different number 
of training shots that much. 

All these observations demonstrate that in the context of robust MAML, the conventional MAML practice, where the number of training shots matches with the number of test shots, may not be optimal.

\textbf{Increasing the number of training shots.}
From Table~\ref{tab: intrinsic dimension full}, we also observe that 
as increasing the number of training shots, the intrinsic dimension of clean representations also increases for both
plain MAML and robust MAML framework. 
Based on these observations, we propose a simple yet effective way, \emph{i.e.}, increasing the number of training shots,
to mitigate the loss of intrinsic dimension caused by adding adversarial loss. 
Formally, for a $N$-way $K$-shot meta-testing task, our method can be expressed as
\begin{equation}    
\begin{aligned}
\underset{\mathbf{\theta}}{\operatorname{minimize}}\ \frac{1}{T}&\sum_{i=1}^T w^c\!\cdot\!\mathcal{L}_{\mathbf{\theta}_i'}(\mathcal{Q}^c)+w^a\!\cdot\!\mathcal{L}_{\mathbf{\theta}_i'}(\mathcal{Q}^a),\\
\operatorname{subject\ to}\ &\mathbf{\theta}_i'=\underset{\theta}{\operatorname{\arg\min}}\ \mathcal{L}_{\theta}(\tilde{\mathcal{S}}^c_i),
\end{aligned}
\label{equatin: its-maml}
\end{equation}
where $\tilde{\mathcal{S}}^c_i=\{s_1, s_2, \cdots, s_{\tilde{K}}\}$ denotes the support set for the $i$-th task during meta-training. 
In our method, the number of support images (\emph{i.e.}, training shot number) $\tilde{K}$ for each task 
at meta-training stage is set to be larger than the number of support images $K$ used in meta-testing.
For example, if we deal with the 5-way 1-shot setting, 
$\tilde{K}$ is set to be larger than 1.
The $w^c$ and $w^a$ are the weights of clean loss $\mathcal{L}_{\mathbf{\theta}_i'}(\mathcal{Q}^c)$
and adversarial loss $\mathcal{L}_{\mathbf{\theta}_i'}(\mathcal{Q}^a)$, respectively. 
Empirically, we find that $w^c=w^a=1$ is sufficient to achieve a good trade-off between clean accuracy and robustness.

Though simple, we empirically find that our method can remarkably improve the clean accuracy of the model without much loss of robustness, thus yielding an accurate yet robust model.
As shown in Fig.~\ref{fig: acc of ITS-MAML under different test shots},
with increasing the number of training shots, 
the clean accuracy steadily increases before reaching a bound. 
For the robustness, it only slightly decreases but still remains high,
which matches with the phenomenon we observed in Table~\ref{tab: intrinsic dimension full}
(\emph{i.e.}, the intrinsic dimension of adversarial noise is quite low and insensitive to the number of training shots).

\section{Experiments}
\label{sec:experiments}

\subsection{Setup}
\textbf{Dataset.}\ \ We conduct experiments on three widely-used few-shot learning benchmarks, \textit{i.e.}, miniImageNet~\cite{vinyals2016matching}, CIFAR-FS~\cite{bertinetto2018meta}, and Omniglot~\cite{lake2015human}. The \textit{miniImageNet} contains 100 classes with 600 samples in each class. The whole dataset is split into 64, 16 and 20 classes for training, validation and testing respectively. We adopt the training set for meta-training, and randomly select 2000 unseen tasks from the testing set for meta-testing. Each image is downsized to $84\times 84\times 3$ in our experiments.
\textit{CIFAR-FS} has the same dataset splitting as miniImageNet, \emph{i.e.}, 64, 16 and 20 classes for training, validation and testing respectively, with each class containing 600 images. We also adopt the training set for meta-training, and randomly select 4000 unseen tasks from the testing set for meta-testing. Each image is resized to $32\times 32\times 3$.
\textit{Omniglot} includes handwritten characters from 50 different alphabets, with a total of 1028 classes of training data and 423 classes of testing data. We randomly select 2000 unseen tasks from the testing data for meta-testing. Each image has a size of $28\times 28\times 1$.


\textbf{Training.} We verify our method based on two kinds of architectures, \emph{i.e.}, a four-layer convolutional neural network as in~\cite{wang2021fast} and a ResNet-12~\cite{he2016deep}.
All the models in our experiments are trained for 12 epochs unless specified. During meta-training, each episode consists of 4 randomly selected tasks. For 5-way 1-shot meta-testing tasks, the number of support images per class in each task of meta-training is 1 for previous methods, and 2 for ITS-MAML. For 5-way 5-shot meta-testing tasks, the number of support images per class in each task of meta-training is 5 for previous methods, and 6 for ITS-MAML. The number of query images per class is 15 for all methods on miniImageNet and CIFAR-FS. On Omniglot, since each class only contains 20 samples, the number of query images per class is thus set to 9 for 5-way 1-shot meta-testing tasks and 5 for 5-way 5-shot meta-testing tasks to avoid data repetition in a task.
The learning rate is set to 0.01 for fine-tuning, and 0.001 for the meta-update. Following~\cite{wang2021fast}, the number of fine-tuning steps is set to 5 for meta-training, and 10 for meta-testing for all methods unless specified. As~\cite{wang2021fast} showed that adopting the FGSM attack~\cite{goodfellow2014explaining} instead of the PGD attack~\cite{madry2017towards} during meta-training can improve the training efficiency without significantly affecting the model performance, we also adopt the FGSM attack~\cite{goodfellow2014explaining} as the training attack. The training attack power $\epsilon$ is set to 2 for miniImageNet, and 10 for CIFAR-FS and Omniglot. 
We evaluate our method under different kinds of attacks for testing. The testing attack power is 2 for miniImageNet, 8 for CIFAR-FS and 10 for Omniglot unless specified.
For a fair comparison, we set $w^c:w^a=1:1$ in Eq.~\ref{equatin: its-maml} for all methods unless specified.
\begin{table*}[t]
\centering
\caption{Accuracy of meta-models trained by different methods under different types of attacks on miniImageNet. The best results are marked in bold.}
\label{tab: main result on miniimagenet}
\vspace{+1.5mm}
\setlength{\tabcolsep}{1mm}{
\begin{tabular}{cccccclcccc}
\toprule[1pt]
\multirow{2}{*}{Method} & \multirow{2}{*}{Model} & \multicolumn{4}{c}{5-way 1-shot accuracy (\%)} &  & \multicolumn{4}{c}{5-way 5-shot accuracy (\%)} \\ \cmidrule[0.3pt]{3-6} \cmidrule[0.3pt]{8-11} 
                        &                        & Clean  & FGSM  & PGD   & CW &  & Clean  & FGSM  & PGD  & CW  \\ \midrule[1pt]
MAML                    & 4-layer CNN            & \textbf{45.00} & 3.71    & 0.60 &  0.24  &  &   \textbf{58.64}     &  7.36 & 1.72 &   2.14     \\
ADML                    & 4-layer CNN            & 37.77  &  29.96     & 27.15 & 26.79   &  & 56.02       &  41.96     & 35.90  &   35.66    \\
AQ                      & 4-layer CNN            & 34.69  &   30.53    & 28.08 & 26.20   &  &  52.66      &   42.82   &  37.21    &    37.84    \\
R-MAML                  & 4-layer CNN            & 37.16  &    29.95   & 27.87 &  26.13  &  &   55.85     &   42.63  &  36.30    &    34.93    \\ \midrule[0.3pt]
\textbf{ITS-MAML (2-shot)}       & 4-layer CNN            & 41.68  &   \textbf{31.74}    & \textbf{28.12} & \textbf{27.84}   &  &  53.37     &  \textbf{43.68}    & \textbf{37.38}    & \textbf{38.22}        \\
\textbf{ITS-MAML (6-shot)}       & 4-layer CNN            & \textbf{43.82}  & 30.60    & 26.71 & 26.34   &  &  \textbf{57.03}    &  42.81     & 35.84    & 35.40       \\ \midrule[1pt]
MAML                    & ResNet-12              &   \textbf{53.27}     &    8.00   &    3.00   & 2.46    &  &    \textbf{70.18}    & 19.22    &   7.18   &   10.28    \\
ADML                    & ResNet-12              &  51.32      &    33.53   & 31.30      & 29.96   &  &  66.08     &   45.30   &  42.10   &   44.73    \\
AQ                      & ResNet-12              &  49.36      &    33.91   &  32.40     & 33.06   &  &    64.85    &  47.82    & \textbf{45.46}     &  45.98       \\
R-MAML                  & ResNet-12              &   50.71     &  33.08     &  31.14     &  30.28  &  & 67.84   &  47.11    &   44.79   &  45.77    \\ \midrule[0.3pt]
\textbf{ITS-MAML (2-shot)}       & ResNet-12              &   52.78     &  \textbf{34.26}     &  \textbf{32.57}     & \textbf{33.18}   &  &  67.07      &  \textbf{47.84}     & 45.12     &  \textbf{46.20}       \\
\textbf{ITS-MAML (6-shot)}       & ResNet-12              &   \textbf{53.00}   &  33.14     &  30.88     & 30.50   &  & \textbf{68.72}    & 46.96     & 43.73    &  45.08     \\ \bottomrule[1pt]
\end{tabular}}
\end{table*}

\begin{table*}[t]
\centering
\caption{Accuracy of meta-models trained by different methods under different types of attacks on CIFAR-FS. A 4-layer CNN is adopted for all methods.}
\label{tab: main result on cifar-fs}
\vspace{+1mm}
\setlength{\tabcolsep}{1mm}{
\begin{tabular}{ccccclcccc}
\toprule[1pt]
\multirow{2}{*}{Method} & \multicolumn{4}{c}{5-way 1-shot accuracy (\%)} &  & \multicolumn{4}{c}{5-way 5-shot accuracy (\%)} \\ \cmidrule[0.3pt]{2-5} \cmidrule[0.3pt]{7-10} 
                        & Clean  & FGSM  & PGD  & CW  &  & Clean  & FGSM  & PGD  & CW  \\ \midrule[1pt]
MAML                    &  \textbf{49.93}      &  9.57     &   0.15   &  0.08   &  &  \textbf{65.63}      &  18.18     &   0.70  &   1.22   \\
ADML                    &   40.41     &  36.67     &  26.05    &  25.79   &  &  56.52      &  49.63     &   32.08   &  32.86     \\
AQ                      &   32.08     &    30.69   &  26.49    &  23.08   &  &   51.40     &   48.19    &  35.08    &   33.90     \\
R-MAML                  &   39.14     &   35.79    &   26.51   &   25.77  &  & 56.09       &  50.78     &  32.67    &  34.38      \\ \midrule[0.3pt]
\textbf{ITS-MAML (2-shot)}       &   42.55     &   \textbf{38.52}    &   \textbf{26.96}   &   \textbf{27.60}  &  &  53.91      &   \textbf{51.46}    &  \textbf{36.47}    & \textbf{37.21}      \\
\textbf{ITS-MAML (6-shot)}       &   \textbf{43.12}     &   38.38    &   25.37   &   26.24  &  &  \textbf{57.12}      &   51.07    &  33.62    & 34.05      \\ \bottomrule[1pt]
\end{tabular}}
\end{table*}

\begin{table*}[t]
\centering
\caption{Accuracy of meta-models trained by different methods under different types of attacks on Omniglot. A 4-layer CNN is adopted for all methods.}
\label{tab: main result on omniglot}
\vspace{+1mm}
\setlength{\tabcolsep}{1mm}{
\begin{tabular}{ccccclcccc}
\toprule[1pt]
\multirow{2}{*}{Method} & \multicolumn{4}{c}{5-way 1-shot accuracy (\%)} &  & \multicolumn{4}{c}{5-way 5-shot accuracy (\%)} \\ \cmidrule[0.3pt]{2-5} \cmidrule[0.3pt]{7-10} 
                        & Clean  & FGSM  & PGD  & CW   &  & Clean  & FGSM  & PGD  & CW   \\ \midrule[1pt]
MAML                    &   93.02     &   65.20    &  22.72    &  26.64   &  &  97.78      &   91.75    &  60.25   &  54.39     \\
ADML                    &  90.58      & 89.84      &  78.27    &  77.19   &  &  97.46      &  96.53     &   91.06   &   \textbf{91.50}     \\
AQ                      &  90.09      &  88.72     &   81.69   &   80.92  &  &  97.02      &  96.44     &  91.25    &  90.11      \\
R-MAML                  &   89.60     &  87.65     &  77.44    &  77.57   &  & 97.12       &   96.09    &   90.28   &  89.72      \\ \midrule[0.3pt]
\textbf{ITS-MAML (2-shot)}       &  93.56      &   \textbf{91.43}    &  \textbf{85.45}    &  \textbf{83.68}   &  &  97.02      &  \textbf{96.80}     &  \textbf{91.87}    &  91.46       \\
\textbf{ITS-MAML (6-shot)}       &  \textbf{94.23}      &   86.89    &  79.90    & 77.34   &  &  \textbf{98.19}      &  96.19     &  90.77    & 88.52       \\ \bottomrule[1pt]
\end{tabular}}
\end{table*}

\begin{figure*}[t]
    \centering
    \includegraphics[width=0.8\linewidth]{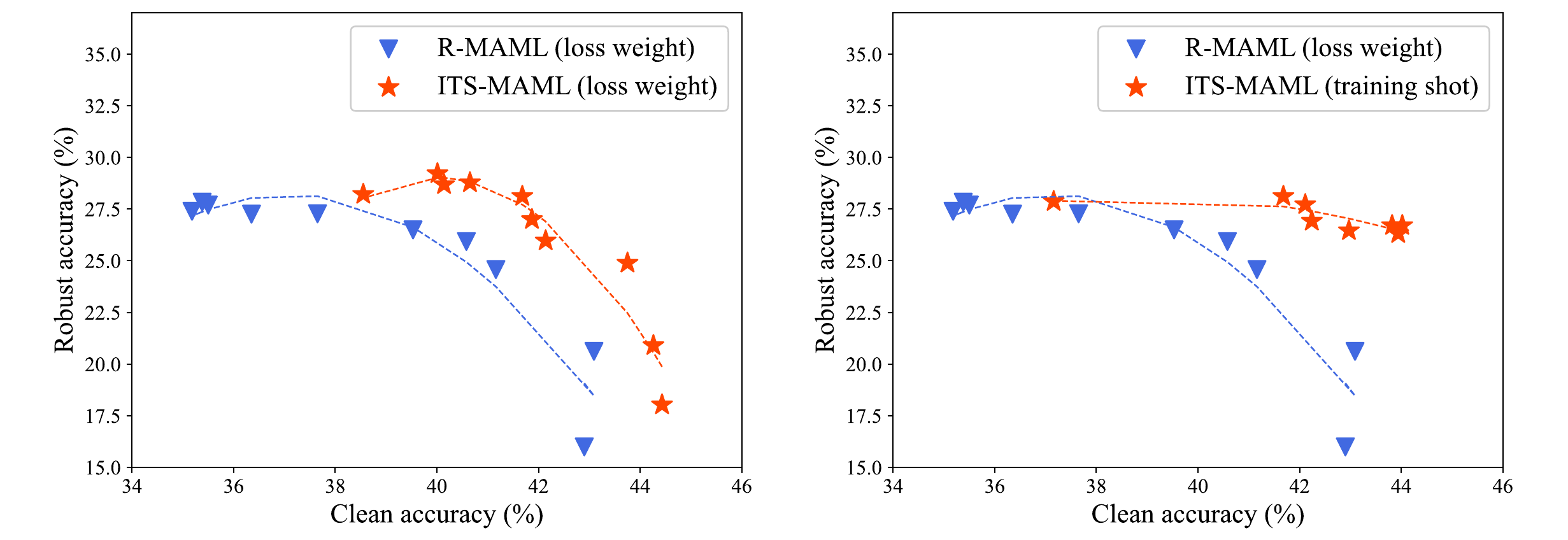}
    \caption{Clean accuracy \emph{vs.} robust accuracy. \textbf{Left}: The results are obtained by varying the value of $w^c\!:\!w^a$ from $1\!:\!0.1$ to $1\!:\!5$ for both R-MAML and ITS-MAML ($\tilde{K}\!=\!2$). \textbf{Right}: The results are obtained by varying the value of $w^c\!:\!w^a$ for R-MAML, and by varying the training shot number $\tilde{K}$ from 1 to 7 for ITS-MAML ($w^c\!:\!w^a\!=\!1\!:\!1$).}
    \label{fig:trade_off}
\end{figure*}

\begin{figure*}
    \centering
    \includegraphics[width=0.9\linewidth]{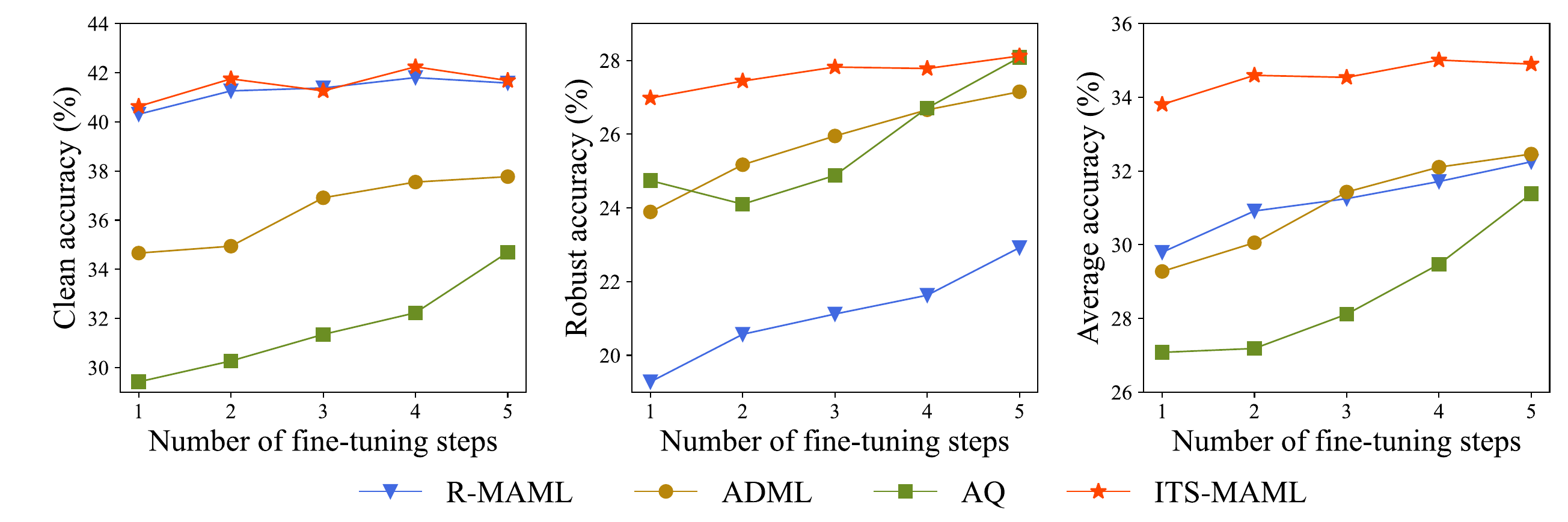}
    \caption{Accuracy of models trained by ADML, AQ, R-MAML ($w^c\!:\!w^a\!=\!1\!:\!0.2$ as in~\cite{wang2021fast}) and ITS-MAML with different number of fine-tuning steps during meta-training. \textbf{Left}: clean accuracy. \textbf{Middle}: Robust accuracy, which is computed with a 10-step PGD attack with $\epsilon=2$. \textbf{Right}: Average accuracy, which denotes the average value of clean accuracy and robust accuracy.
    }
    \label{fig: efficiency}
\end{figure*}

\subsection{Comparisons with previous methods}
We evaluate the performance of our proposed method (denoted as ``ITS-MAML'') under 5-way 1-shot and 5-way 5-shot settings, and compare our method with plain MAML and previous typical robust MAML methods, \emph{i.e.}, AQ, ADML and R-MAML. We demonstrate the experimental results on three benchmarks in Tables~\ref{tab: main result on miniimagenet},~\ref{tab: main result on cifar-fs} and~\ref{tab: main result on omniglot}, respectively. In those tables, we show the accuracy on clean images (``Clean'') and robust accuracy under different types of attacks, \emph{i.e.}, FGSM~\cite{goodfellow2014explaining}, PGD~\cite{madry2017towards}, and CW~\cite{carlini2017towards}. 
The accuracy under different types of attacks reflects the model's robustness level.
Based on the experiment results, we have three observations:
\textbf{1)} Compared to MAML, all the robust MAML methods (including ours) significantly improve the model's robustness. For example, for 4-layer CNN on 5-way 1-shot task of miniImageNet, compared to MAML baseline, AQ, ADML, R-MAML and ITS-MAML (2-shot) improve the accuracy under PGD attack by about 27\%, 27\%, 27\% and 28\%, respectively. These results verify that introducing adversarial loss into MAML contributes to the model's robustness.
\textbf{2)} By comparing previous robust MAML methods (\emph{e.g.}, ADML and R-MAML), we find that generally they achieve comparable results for both clean accuracy and robustness, indicating whether applying the adversarial loss to the fine-tuning procedure doesn't bring a huge difference. For example, for 4-layer CNN on 5-way 1-shot task of miniImageNet, the clean accuracy for ADML and R-MAML is 37.77\% and 37.16\% respectively, while the robust accuracy is 27.15\% and 27.87\% (under PGD attack). Generally, AQ performs worse than ADML and R-MAML on clean accuracy, which indicates imposing clean loss for meta-update may be necessary for a high clean accuracy. 
\textbf{3)} Compared to previous robust MAML methods, our increasing training shot number strategy (ITS-MAML) 
remarkably improves the clean accuracy while maintaining good robustness (sometimes the robustness is even better than previous methods).
For example, on 5-way 1-shot task on CIFAR-FS, in terms of clean accuracy, ITS-MAML (2-shot) outperforms AQ and R-MAML by more than 10\% and 3\%. If we increase the training shot number (\emph{i.e.}, from 2-shot to 6-shot), we observe that ITS-MAML (6-shot) further improves the clean accuracy by around 1\%. The robust accuracy decreases slightly, but still remains at a high level, which is competitive among previous robust MAML methods. 
For 5-way 5-shot task, we observe the similar trend. Thus, if we care more about the model's robustness, we may lower the number of training shots. However, if we want to achieve a robust yet accurate model, increasing the training shot number is a good choice.

\subsection{Ablation Study and Analysis}
\label{subsec: better trade-off}

\textbf{Better trade-off between clean accuracy and robustness.}
R-MAML~\cite{wang2021fast} demonstrated superior clean accuracy compared to AQ and ADML,
when the learning rate of clean loss is set to 5 times the learning rate of adversarial loss,
\emph{i.e.}, the ratio $w^c\!:\!w^a$ is set to $1\!:\!0.2$ (see Eq.~\ref{equatin: its-maml}).
We find that when this ratio is set to be consistent with ADML and ITS-MAML, \emph{i.e.}, $1\!:\!1$, the high clean accuracy of R-MAML no longer exists (see Fig.~\ref{fig:motivation_trade_off_figure}).
It implies that there exists a trade-off between clean accuracy and robustness with the change of this ratio.
We compare our ITS-MAML with R-MAML under different ratios of $w^c\!:\!w^a$ and show corresponding clean accuracy and robustness in Fig.~\ref{fig:trade_off}. To demonstrate the effectiveness of increasing training shot number, we also show the results of ITS-MAML under different number of training shots (keeping $w^c\!:\!w^a\!=\!1\!:\!1$) in Fig.~\ref{fig:trade_off}.
From Fig.~\ref{fig:trade_off} (left), we observe that 
the larger the ratio of $w^c\!:\!w^a$, the higher the clean accuracy and the lower the robust accuracy of the model.
Compared with R-MAML, our ITS-MAML achieves obviously better trade-off (as shown, the curve of ITS-MAML is above that of R-MAML). 
In addition, Fig.~\ref{fig:trade_off} (right) demonstrates that with increasing the number of training shots, the clean accuracy of ITS-MAML steadily increases before reaching a bound, while the robustness remains relatively stable at a high level, which further verifies the much better trade-off achieved by our method. 



\begin{wrapfigure}[18]{r}{0.45\linewidth}
    \begin{center}
        \includegraphics[width=0.95\linewidth]{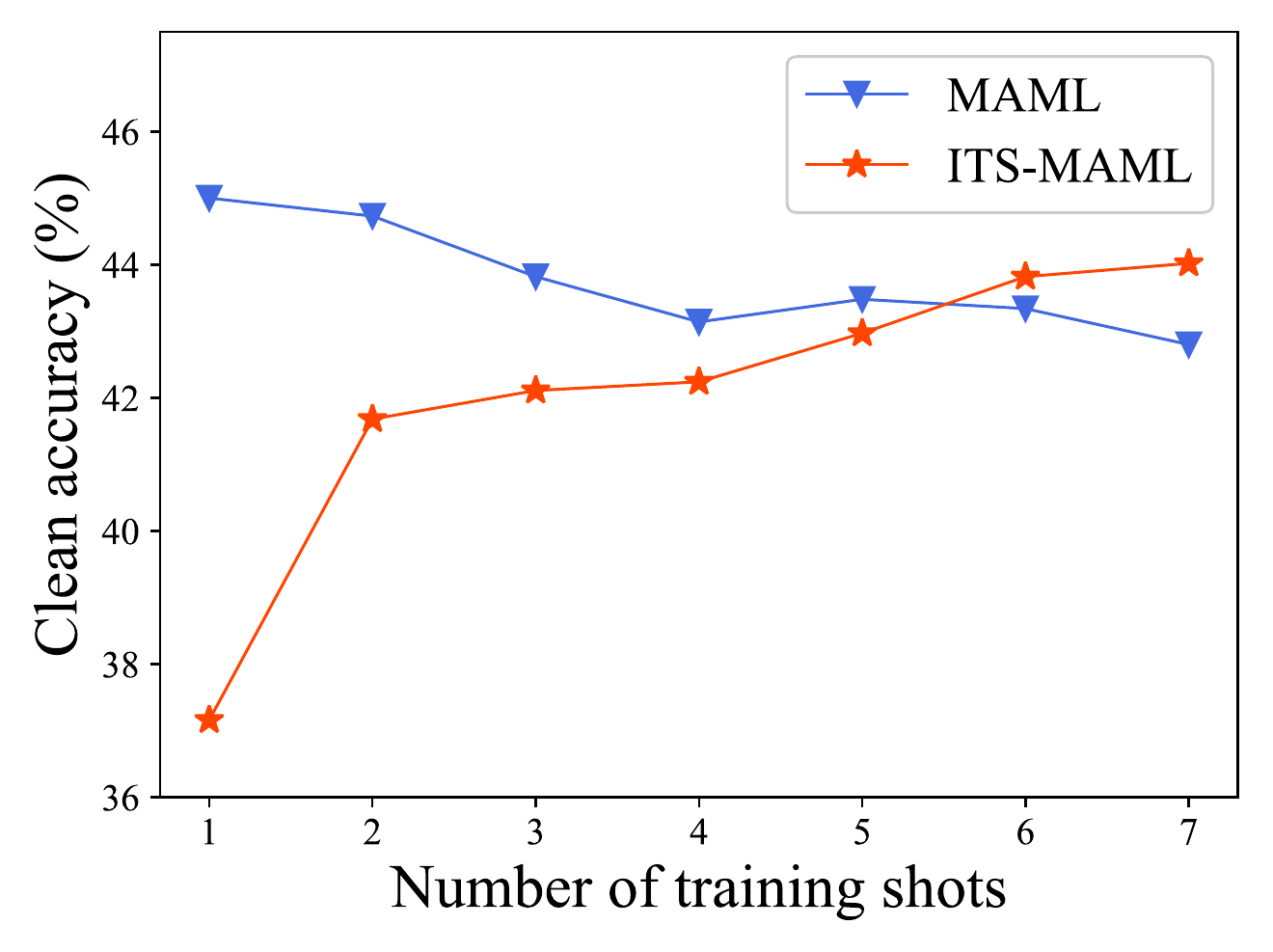}
    \end{center}
    \caption{Clean accuracy for 1-shot testing of models trained by MAML and ITS-MAML with different number of training shots.}
    \label{fig: clean accuracy of maml and its-maml}
\end{wrapfigure}

\label{subsec: better efficiency}
\textbf{More efficient training.}
The number of fine-tuning steps (\emph{i.e.}, $M$ in Eq.~\ref{equation: fine-tuning steps}) during meta-training affects the training efficiency of the model. Larger $M$ leads to higher computation costs and a lower training efficiency. We expect that ITS-MAML can achieve good performance even if the number of fine-tuning steps $M$ is reduced during training, thus improving the training efficiency. To this end, we train ADML, AQ, R-MAML ($w^c\!:\!w^a\!=\!1\!:\!0.2$ as in~\cite{wang2021fast}) 
and ITS-MAML models with $M$ ranging from 1 to 5. We then evaluate the 5-way 1-shot accuracy of the models on miniImageNet. The results are shown in Fig.~\ref{fig: efficiency}. We observe that previous methods may suffer from an obvious performance degradation (either clean accuracy or robustness) with reducing $M$. Compared to previous methods, our ITS-MAML is less sensitive to the number of fine-tuning steps during meta-training. This allows for a reduced number of fine-tuning steps to improve the training efficiency.
\textbf{Effect of number of shots on MAML and Robust MAML.}
\label{sec:exp-intrinsic-dim}
To show the different effect of number of shots on plain MAML and robust MAML, we consider a 5-way 1-shot task on miniImageNet and 
demonstrate the clean accuracy curve of plain MAML and robust MAML with respect to different number of training shots in Fig.~\ref{fig: clean accuracy of maml and its-maml}.
As we expect, the clean accuracy decreases as we increase the training shot number for plain MAML, while for robust MAML, 
the trend is different, \emph{i.e.}, the clean accuracy steadily increases before reaching a bound.
Such a phenomenon is consistent with what we can derive from the trend of intrinsic dimension of clean features (see Table~\ref{tab: intrinsic dimension full}).
For the 5-way 1-shot setting, as shown in Table~\ref{tab: intrinsic dimension full}, the suitable intrinsic dimension is about 80.
In plain MAML, if we adopt a larger number of training shots, the resulting intrinsic dimension will be larger than 80, 
\emph{e.g.}, for 2-shot, the intrinsic dimension is 103, which is much larger than 80 (1-shot).
A larger intrinsic dimension may not be affordable when we fine-tune our meta-model at the testing stage (1-shot), resulting in 
varying degrees of overfitting.
However, for robust MAML, as discussed before, increasing the number of training shots may mitigate the capacity loss of representations, thus improving the accuracy at the testing stage.

\section{Conclusion}

In this paper, we observe that introducing adversarial loss into MAML framework reduces the intrinsic dimension of features, which results in a lower capacity of representations of clean samples. 
Based on this observation, we propose a simple yet effective strategy, \emph{i.e.}, increasing the number of
training shots, to mitigate the loss of intrinsic dimension caused by introducing adversarial loss.
Extensive experiments on few-shot learning benchmarks demonstrate that compared to previous robust MAML methods, our method can achieve superior clean accuracy while maintaining high-level robustness. Further, empirical studies show
that our method achieves a better trade-off between clean accuracy and robustness, and has better training efficiency.
We hope our new perspective may inspire future research in the field of robust meta-learning.

\section*{Acknowledgments}
This work was supported in part by Zhongguancun Laboratory, Beijing, P.R.China.




\end{document}